%% file: main.tex
\documentclass[runningheads]{llncs}
\makeatletter
\newcommand{\printfont}{Encoding: \f@encoding{}, Family: \f@family{}, Series: \f@series{}, Shape: \f@shape{}}
\makeatother
 
\usepackage{eccv}

\usepackage{eccvabbrv}

\usepackage{graphicx}
\usepackage{booktabs}
\usepackage{tabularx}
\usepackage{multirow}
\usepackage{indentfirst}
\usepackage{float}
\usepackage{caption}
\input{preamble}
\usepackage[accsupp]{axessibility}  

\usepackage{hyperref}

\usepackage{orcidlink}

\def\etal{\emph{et al}\onedot}

\begin{document}
\title{From Pixels to Objects: A Hierarchical Approach for Part and Object Segmentation Using Local and Global Aggregation}

\titlerunning{LGFormer for Part and Object Segmentation}


\author{Yunfei Xie\inst{1}\orcidlink{0009-0000-8853-1751} \and
Cihang Xie\inst{2}\orcidlink{0000-0003-1243-8045} \and
Alan Yuille\inst{3}\orcidlink{0000-0001-5207-9249} \and 
Jieru Mei\inst{3}\orcidlink{0000-0002-4710-9463}}

\authorrunning{Y. Xie et al.}


\institute{Huazhong University of Science and Technology
\and
UC Santa Cruz\\
\and 
Johns Hopkins University
}

\maketitle

\input{sec/0_abstract}    
\input{sec/1_intro}

\input{sec/2_relatedworks}
\input{sec/3_method}

\input{sec/4_experiments}
\input{sec/5_conclusion}

\section*{Acknowledgements}

This work was supported by ONR N00014-21-1-2690.

{
    \small
    \bibliographystyle{splncs04}
    \bibliography{main}
}

\renewcommand{\thesection}{\Alph{section}}
\setcounter{section}{0}

\end{document}

%% file: preamble.tex
\usepackage[dvipsnames]{xcolor}

\def\etal{\emph{et al}\onedot}

\newcommand{\pixel}{pixels\xspace}
\renewcommand{\sp}{superpixels\xspace}
\newcommand{\group}{groups\xspace}

\newcommand{\modelnamefull}{Local Global Transformer\xspace}
\newcommand{\modelname}{LGFormer\xspace}
\newcommand{\upsamplename}{association-aware upsampling\xspace}
\newcommand{\sca}{SCA\xspace}
\newcommand{\scafull}{Superpixel Context Aggregation\xspace}
\newcommand{\gca}{GCA\xspace}
\newcommand{\gcafull}{Group Context Aggregation\xspace}

%% file: sec/0_abstract.tex
\begin{abstract}

In this paper, we introduce a hierarchical transformer-based model designed for sophisticated image segmentation tasks, effectively bridging the granularity of part segmentation with the comprehensive scope of object segmentation. At the heart of our approach is a multi-level representation strategy, which systematically advances from individual pixels to superpixels, and ultimately to cohesive group formations. This architecture is underpinned by two pivotal aggregation strategies: local aggregation and global aggregation. Local aggregation is employed to form superpixels, leveraging the inherent redundancy of the image data to produce segments closely aligned with specific parts of the object, guided by object-level supervision. In contrast, global aggregation interlinks these superpixels, organizing them into larger groups that correlate with entire objects and benefit from part-level supervision. This dual aggregation framework ensures a versatile adaptation to varying supervision inputs while maintaining computational efficiency.

Our methodology notably improves the balance between adaptability across different supervision modalities and computational manageability, culminating in significant enhancement in segmentation performance. When tested on the PartImageNet dataset, our model achieves a substantial increase, outperforming the previous state-of-the-art by 2.8\% and 0.8\% in mIoU scores for part and object segmentation, respectively. Similarly, on the Pascal Part dataset, it records performance enhancements of 1.5\% and 2.0\% for part and object segmentation, respectively.
  \keywords{Semantic segmentation \and Superpixels}

\end{abstract}

%% file: sec/1_intro.tex
\section{Introduction}
\label{sec:lg_intro}

\begin{figure*}[htbp]
  \centering
  \includegraphics[page=1, width=0.8\textwidth]{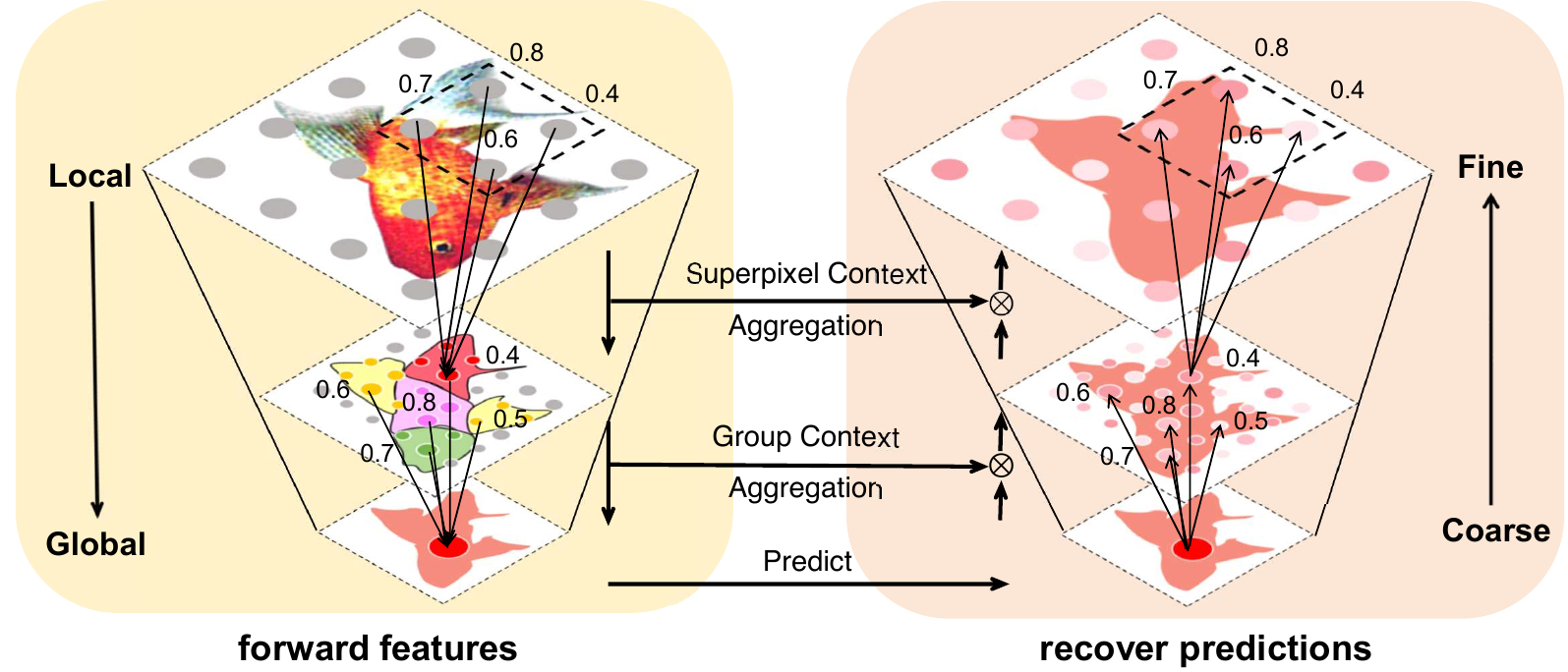}
  \caption{
\textbf{Conceptual illustrations of \modelname.} On the left, \modelname follows a hierarchical aggregation pathway, elevating features from pixels to parts to objects. The right figure shows the model's ability to progressively restore segmentation predictions from the object level to the original image resolution.
}
\label{fig:lg_teaser}
\end{figure*}

Joint part and object segmentation presents a formidable challenge that involves simultaneously performing holistic object segmentation and detailed part segmentation. Although current methods \cite{wang2015joint,singh2022float,he2023compositor,zhang2024dictionary} have shown effectiveness in segmenting object parts, there often remains a gap in achieving simultaneous object-level segmentation, as well as constraints related to computational efficiency. In particular, many approaches \cite{singh2022float,michieli2020gmnet,zhao2019multi,peng2024learning,peng2024dspart} are tailored primarily for part segmentation and do not effectively address the dual task of object segmentation. Additionally, prevalent approaches \cite{wang2015joint} may employ separate specialized computational architectures for each segmentation task, thereby substantially increasing the computational overhead.


The underlying challenge stems from the inherently conflicting goals of part and object segmentation. Object segmentation necessitates the integration of broad features across an object to ensure a cohesive representation, in contrast to part segmentation, which requires distinguishing between the smaller, detailed features within the object. Moreover, these segmentation tasks differ fundamentally in their spatial emphasis: object segmentation is enhanced by a broader, global perspective which aids in object recognition, whereas part segmentation focuses more narrowly on local details essential for accurate delineation of component boundaries \cite{wang2015joint}. This inherent tension demands a strategy that can reconcile global coherence with detailed local recognition, highlighting the urgency for innovative solutions capable of efficiently bridging the conflicting demands of holistic object and intricate part segmentation within a cohesive framework.


To address these complexities, we introduce \textbf{\modelnamefull} (\modelname), a model inspired by the hierarchical structure of human visual perception, which starts with recognizing smaller components and their spatial relationships before synthesizing these elements into a holistic understanding of the object \cite{lake2015human}. \modelname is designed to innovatively manage the simultaneous segmentation of detailed parts and entire objects, thereby addressing the challenge from a foundational level.

\textbf{Hierarchical Representation.} At the core of our approach is the hierarchical organization of visual data, which simulates the natural progression from discrete pixels to complex object representations. This system organizes \pixel into \sp, and these \sp into \group through advanced cross-attention mechanisms, enabling \modelname to capture multi-scale information adeptly. This structured arrangement facilitates concurrent detailed part segmentation and comprehensive object segmentation, maintaining precision and fidelity across varying scales.

\textbf{Association-Aware Upsampling.} Building on the structured hierarchy of \pixel, \sp, and \group, \modelname incorporates an innovative upsampling technique, termed \upsamplename, engineered to preserve and accurately convey detailed spatial information through the hierarchical layers back to the original image scale. By utilizing attention scores from cross-scale interactions, this method preserves the integrity of fine details more effectively than traditional upsampling approaches, thereby enhancing both precision and detail retention. This association-aware upsampling serves as a critical component, ensuring the preservation of multi-scale segmentation insights in the final high-resolution output.

\textbf{Unified Multi-Task Framework.} The foundation of \modelname is a unified architectural approach that obviates the need for separate frameworks for part and object segmentation. This integrated structure not only simplifies the segmentation workflow but also boosts the model’s efficiency and interpretability. By employing a single, cohesive framework, \modelname adeptly shifts between the local and global segmentation demands, marking a substantial advancement in replicating human-like perception in visual segmentation tasks.

Thus, \modelname stands as a pioneering solution in the domain of joint part and object segmentation, offering a scalable and efficient approach that adeptly balances meticulous detail segmentation with global context comprehension, tackling both local and global redundancies with unmatched precision.

We empirically validate the superior performance of \modelname on the benchmark datasets PartImageNet~\cite{he2022partimagenet} and Pascal-Part~\cite{Everingham2010ThePV}. The model demonstrates its capability in generating high-quality semantic parts, substantially enhancing object segmentation. Our evaluations show that \modelname achieves mIoU scores of 67.4\% and 79.8\% for part and object segmentation, respectively, surpassing the previous state-of-the-art model, Compositer \cite{he2023compositor}, by 2.9\% and 0.9\%. Remarkably, on the Pascal-Part dataset, \modelname exceeded Compositer by 1.5\% and 2.0\% in part and object mIoU, respectively.


Further examination underscores the capability of \modelname's hierarchical architecture to engender semantic understanding autonomously, illuminating a process that is both visually interpretable and inherently explainable. Notably, this architecture enables the derivation of \group from \sp without explicit object-level supervision, and similarly allows \sp to form solely from object supervision. These dynamics are comprehensively detailed in \cref{fig:emerge}, showcasing the intrinsic adaptability and intelligence of our hierarchical representation in capturing complex semantic relationships.


\begin{enumerate}
\item We introduce a novel hierarchical representation that emulates the functioning of the human visual system, effectively reducing the computational complexity of Vision Transformers. This approach distinctively addresses both local and global redundancies, enhancing processing efficiency.
\item Leveraging this hierarchical structure, we develop an innovative upsampling technique termed \upsamplename. This method successfully overcomes the blurring commonly associated with existing upsampling strategies, thereby preserving the fidelity of fine details across various segmentation tasks.
\item We utilize our hierarchical framework to manage both part and object segmentation within a unified model, \modelname, demonstrating our approach's versatility and efficiency.
\end{enumerate}

%% file: sec/2_relatedworks.tex
\section{Related Works}
\label{sec:relatedworks}

\noindent \textbf{Bridging Part and Object Segmentation}\quad
Joint learning of objects and parts representation concurrently is an attractive topic and has been actively investigated. Wang \etal \cite{wang2015joint} introduced a novel approach with a dual-channel, fully convolutional network that predicts semantic compositional parts and object potentials at the pixel level, complemented by a fully connected conditional random field for refined predictions. Subsequently, Singh \etal's FloatSeg \cite{singh2022float} framework innovatively involves multiple decoders for objects and part attributes respectively. However, these methodologies share a common drawback: their reliance on multiple encoders or decoders for handling part and object representations, which increases computational demand and complexity in mapping part-to-object relationships.
The most related work to ours is Compositor \cite{he2023compositor}, which proposes a bottom-up strategy to compose embeddings from parts to objects.
However, in Compositor, interactions among pixels, parts, and objects operate completely at a global scale, which is suboptimal due to the high computational cost and the absence of local attention as inherent inductive biases. In contrast, our model leverages the \sp to utilize the local redundancy within the image, yielding segments that align with image parts even under solely object-level supervision, making it a natural fit for part segmentation. Thanks to the boundary preservation from our hierarchy representation, our association-aware upsampling produces sharper predictions.

\noindent \textbf{Overcoming Local and Global Redundancies}\quad  
High-resolution image processing often grapples with redundancies that challenge both resolution management and detail preservation. Traditional techniques employ max pooling to downscale images \cite{krizhevsky2012imagenet,he2015deep,howard2017mobilenet,sandler2018mobilenetv2,mei2020atom,liu2021swin}, albeit at the cost of losing finer details. To counteract this loss, extensive decoders have been introduced to recover detailed information \cite{Long2015CVPR, Chen2018TPAMI, Chen2017ARXIV}. In contrast, the incorporation of superpixels into deep learning frameworks \cite{Jampani2018ssn, yang2020superpixel, locatello2020object, xu2022groupvit, yu2022cmt, zhang2022semantic, yu2022kmeans_transformer, ma2023image, huang2023stvit, mei2023superformer_final, mei2024spformer} presents a more nuanced approach, effectively addressing local image redundancy while conserving boundary integrity.

Yet, while superpixels adeptly handle local redundancy, global redundancy remains a concern. Recent models like GCViT \cite{hatamizadeh2023global}, GroupViT \cite{xu2022groupvit}, and GPViT \cite{yang2023gpvit} have attempted to bridge this gap by incorporating group tokens that facilitate global information exchange. However, due to computational demands, these models typically rely on either coarse patches or a limited number of \group, which restricts their capacity.

\modelname innovates on this front by sequentially advancing from \pixel to \sp, and finally to \group. This progression addresses both local and global redundancies through a hierarchical representation that aligns seamlessly with the joint demands of part and object segmentation. Unlike its predecessors, \modelname harnesses varying levels of feature semantics, enabling a more comprehensive and detail-preserving segmentation approach.

%% file: sec/3_method.tex
\section{Method}
\label{sec:lg_method}

\begin{figure*}[htbp]
  \centering
      \includegraphics[page=1, width=\textwidth]{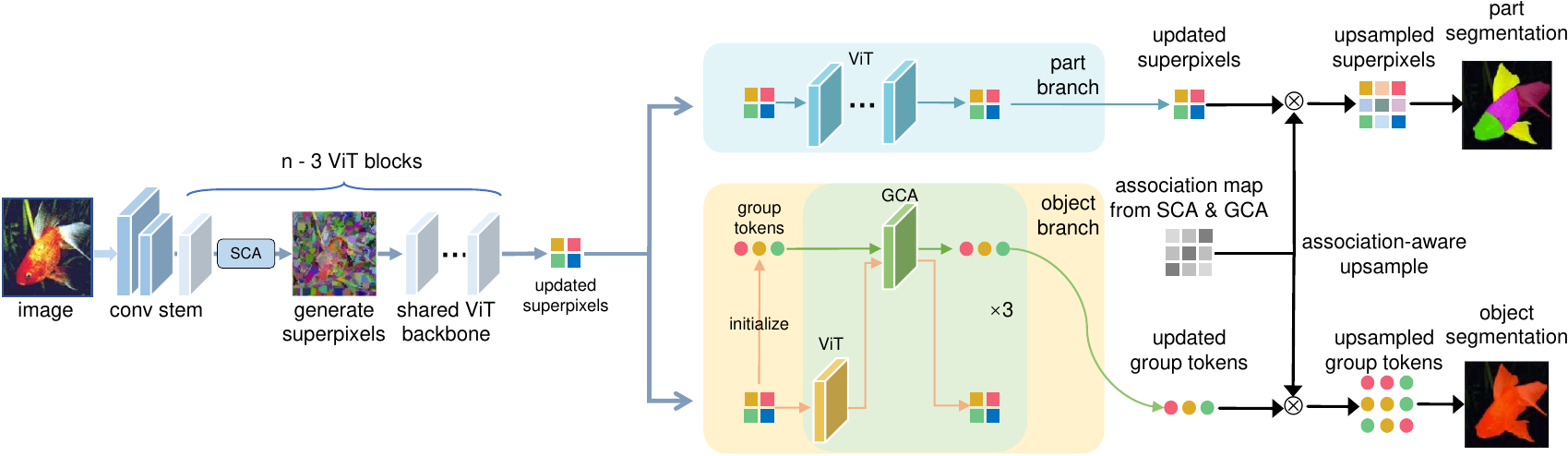}
  \caption{
  \textbf{Overview of \modelname.} Pixel-level features are extracted by a light convolution stem. In the initial ViT stages, these features are refined into part-level superpixels via \scafull (\sca). In deeper ViT layers, \sp are aggregated into object-level groups using \gcafull (\gca).
  }

  \label{fig:lg_arch}
\end{figure*}

\subsection{From Pixels to Objects}
\label{sec:lg_method_1}

\textbf{Pixel Representation.}\quad Given an input image $\mathbf{M} \in \mathbb{R}^{m_h \times m_w \times 3}$, the large spatial dimensions inherently introduce computational complexity. To address this, we use a lightweight convolution stem that downsamples the image and extracts pixel feature maps $\mathbf{I} \in \mathbb{R}^{{i_h} \times {i_w} \times {i_c}}$, thereby reducing initial redundancy and computational load while preserving crucial visual details.

\textbf{Superpixel Representation.}\quad Noting that regions within an object often contain clusters of pixels with redundant information—due to similarity among adjacent pixels—we shift to a superpixel representation $\mathbf{S} \in \mathbb{R}^{{s_h} \times {s_w} \times {s_c}}$. Following \cite{mei2024spformer}, we apply a \scafull (\sca) technique, which aggregates pixels into superpixels by integrating local contextual information. This transition effectively reduces local redundancy and increases the model’s efficiency and explainability by focusing on coherent contextual segments.


\textbf{Group Representation.}\quad Although superpixels efficiently compress local information, they generally fail to capture the global semantics critical for object-level representations. This limitation is due to their focus on localized areas, which misses the broader, holistic view required for recognizing entire objects. To rectify this, we employ a method that groups multiple superpixels into \group $\mathbf{G} \in \mathbb{R}^{{g_n} \times {g_c}}$ using a \gcafull (\gca). This method reduces global redundancy by abstracting similar or repeated part features across the image and simplifies computational demands. More importantly, it enhances global interpretability and overall model performance by forming higher-level abstractions that more accurately reflect object-level semantics.

\begin{figure}[t]
    \centering
    \begin{subfigure}[b]{0.9\textwidth}
        \includegraphics[width=\textwidth]{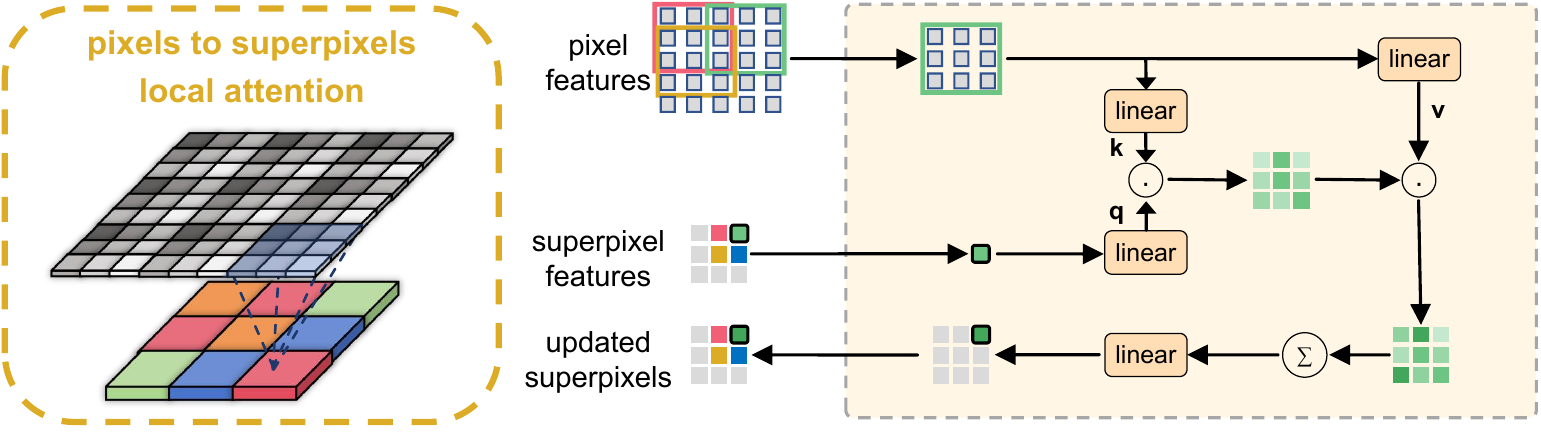}
        \caption{\scafull (\sca)}
        \label{fig:sca}
    \end{subfigure}
    \begin{subfigure}[b]{0.9\textwidth}
        \includegraphics[width=\textwidth]{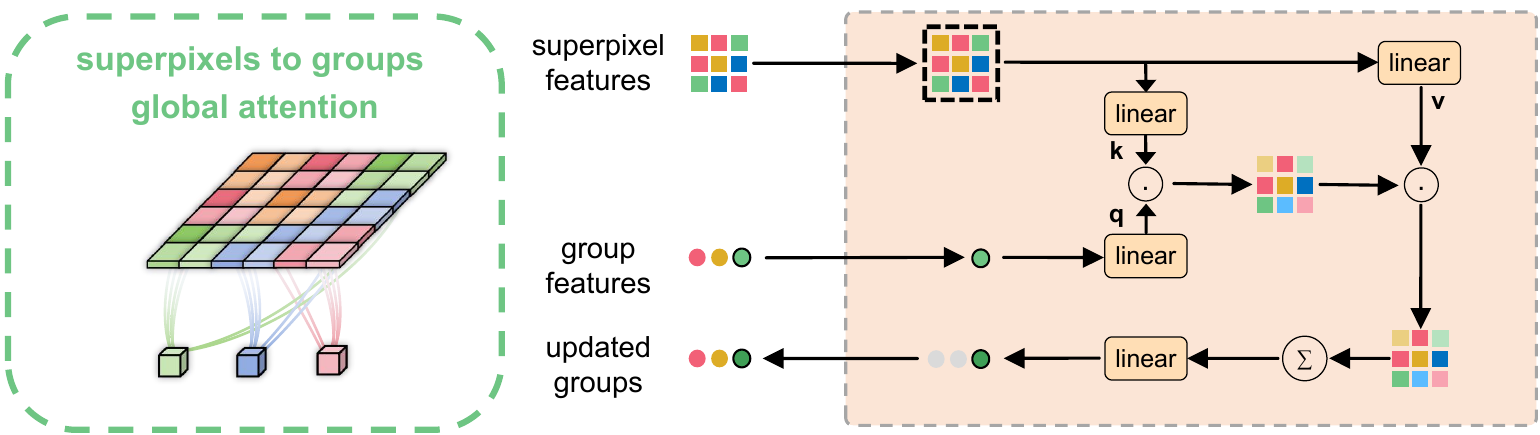}
        \caption{\gcafull (\gca)}
        \label{fig:gca}
    \end{subfigure}
    \caption{
        \textbf{Illustration of interactions and spatial relationships among three hierarchical levels: \pixel, \sp, and \group.} For clarity, the illustration presents a scenario involving only a single superpixel and a single group token. (a) Transitioning from \pixel to \sp involves iterative refinement of \sp on a local scale through \sca. (b) Advancing from \sp to \group, the refinement of \group on a global scale is facilitated by \gca.
    }
    \label{fig:sca_and_gca}
\end{figure}

\textbf{Local and Global Aggregation.}\quad Hierarchical spatial downsampling, a prevalent technique in segmentation models \cite{liu2021swin,chen2018deeplabv2}, typically fails to differentiate adequately between semantic elements, merging distinct features indiscriminately. To mitigate this, we introduce a novel semantic stratification process, by mapping part to \sp and object to \group via \sca and \gca mechanisms, as shown in \cref{fig:sca_and_gca}. This method ensures that semantic redundancies are efficiently managed, preserving essential features while reducing unnecessary information across scales.

In \sca, our approach aims to mitigate local redundancies by efficiently organizing \pixel into \sp. This crucial step not only reduces data complexity but also preserves vital details necessary for precise part segmentation. The process is articulated as:
\begin{equation}
\mathbf{S}^{t}_p = \mathbf{S}^{t-1}_p + \sum_{i \in \mathcal{N}_p} \text{softmax}\left( \mathbf{q}_{\mathbf{S}^{t-1}_p} \cdot \mathbf{k}_{\mathbf{I}^{t-1}_i} \right) \mathbf{v}_{\mathbf{I}^{t-1}_i},
\end{equation}
where \(\mathcal{N}_p\) identifies the pixels adjacent to superpixel \(p\). The objective of this aggregation is to minimize local redundancy through optimized pixel-to-superpixel assignments. The attention mechanism utilizes the softmax function to process the dot products of the query (\(\mathbf{q}\)) and key (\(\mathbf{k}\)) vectors, determining the significance of contributions from each pixel.
These vectors, alongside the value vector (\(\mathbf{v}\)), are derived from linear transformations of the features of superpixels \(\mathbf{S}^{t-1}_p\) and pixels \(\mathbf{I}^{t-1}_i\) from the previous iteration. This permits the model to adaptively focus on and integrate the most relevant pixel data to enhance superpixel features. For comprehensive insights, we direct the reader to SPFormer \cite{mei2024spformer}.

Building upon local context aggregation, \gca employs global cross-attention to iteratively update \group and \sp. This mechanism is made feasible by the reduced number of \group, which lowers computational costs. Additionally, the preceding mitigation of local redundancies by \sca aids in enhancing semantic abstraction at the object level, rendering the aggregation process more efficient.
Each iteration $t$ consists of two critical phases: Superpixel-to-Group (S2G) and Group-to-Superpixel (G2S) cross-attention.

For S2G cross-attention within \gca, group tokens \(\mathbf{G}^{t}_g\) are refined by aggregating information across all superpixels, with an FFN applied to elevate the aggregated features to a higher semantic level:
\begin{equation}
\mathbf{G}^{t}_g = \mathbf{G}^{t-1}_g + \operatorname{FFN}\left(\sum_{p \in \mathcal{P}} \text{softmax}\left( \mathbf{q}_{\mathbf{G}^{t-1}_g} \cdot \mathbf{k}_{\mathbf{S}^{t-1}_p} \right) \mathbf{v}_{\mathbf{S}^{t-1}_p}\right),
\end{equation}
where \(\mathcal{P}\) denotes the set of all superpixels, streamlining the enhancement of group token representations by encompassing comprehensive superpixel insights.

Conversely, the G2S cross-attention phase updates superpixel features \(\mathbf{S}^{t}_p\) by assimilating global \group information, thus ensuring that each superpixel representation benefits from a broader context:
\begin{equation}
\mathbf{S}^{t}_p = \mathbf{S}^{t-1}_p + \operatorname{FFN}\left(\sum_{g \in \mathcal{G}} \text{softmax}\left( \mathbf{q}_{\mathbf{S}^{t-1}_p} \cdot \mathbf{k}_{\mathbf{G}^{t-1}_g} \right) \mathbf{v}_{\mathbf{G}^{t-1}_g} \right),
\end{equation}
where \(\mathcal{G}\) represents the collective set of group tokens, highlighting the interplay between superpixel and group token features for refined segmentation.

These phases facilitate a bidirectional information flow between superpixels and groups, promoting comprehensive semantic integration across both local and global scales. This method ensures that each group captures broader contextual insights while each superpixel receives enriched contextual feedback from the global perspective, thus optimizing the overall segmentation accuracy.

To further enhance global interactions among \group, a ViT block is integrated between the S2G and G2S stages, which further boosts the model's global semantic analysis capabilities:
\begin{equation}
{\mathbf{G}^{t}}' = \operatorname{MHSA}\left(\operatorname{LN}\left(\mathbf{G}^{t}\right)\right) + \mathbf{G}^{t},
\end{equation}
\begin{equation}
\tilde{\mathbf{G}}^{t} = \operatorname{MLP}\left(\operatorname{LN}\left({\mathbf{G}^{t}}'\right)\right) + {\mathbf{G}^{t}}',
\end{equation}
This ViT block integration, utilizing Multi-Head Self-Attention (MHSA) and Layer Norm (LN), enables the model to adeptly navigate complex global interactions at the \group level, further augmenting segmentation precision.

\subsection{Association-Aware Upsampling}

\begin{figure}[tbp]
\centering
    \includegraphics[width=0.8\textwidth]{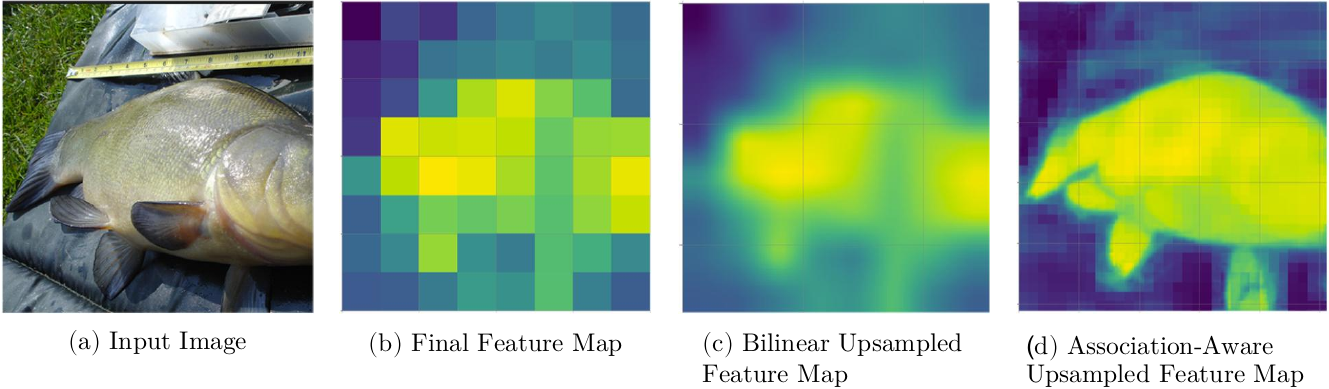}
\caption{
\textbf{Enhanced Detail with Association-Aware Upsampling.} In contrast to the conventional bilinear upsampled feature map, our association-aware upsampled feature map achieves sharper boundary delineation and retains greater semantic detail, which is crucial for detailed segmentation tasks.
}
\label{fig:upsample_vis}
\end{figure}

Unlike traditional hierarchical models with unidirectional information flow from fine to coarse levels \cite{yu2023kmaxdeeplab,yu2022cmtdeeplab}, our approach introduces bidirectional flow. This bilateral hierarchy enables data aggregation from \pixel to \sp, and then to \group, and facilitates detailed reconstruction from coarser to finer scales. This design enhances the model's ability to restore predictions to original resolution more accurately than traditional bilinear upsampling, which often lacks specificity to the data's inherent structure.
    
The core of this methodology lies in the attention scores detailed in \cref{sec:lg_method_1}, which define association matrices elucidating the intricate relationships among the \pixel, \sp, and \group.
Leveraging these matrices enables the systematic upscaling of predictions from the group level ($\mathbf{O}_{\mathbf{G}} \in \mathbb{R}^{g_n \times o_c}$), through the superpixel level ($\mathbf{O}_{\mathbf{S}} \in \mathbb{R}^{s_h \times s_w \times o_c}$), and ultimately, back to the original pixel scale ($\mathbf{O}_{\mathbf{I}} \in \mathbb{R}^{i_h \times i_w \times o_c}$):
\begin{align}
\mathbf{O}_{\mathbf{S}} &= \mathbf{A}_{g \rightarrow p} \cdot \mathbf{O}_{\mathbf{G}}, \\
\mathbf{O}_{\mathbf{I}} &= \mathbf{A}_{p \rightarrow i} \cdot \mathbf{O}_{\mathbf{S}}.
\end{align}
Here, $\mathbf{A}_{g \rightarrow p}$ denotes the association matrix mapping from group token $g$ to superpixel $p$, and $\mathbf{A}_{p \rightarrow i}$ represents the matrix mapping from superpixel $p$ to pixel $i$.
This procedural flow intricately enhances coarse object predictions, incrementally refining them to enrich part shapes at the \sp level and meticulously refine boundaries at the pixel level. Through the bidirectional flow of information, our methodology not only preserves but also enhances the semantic integrity of the upscaled predictions, ensuring fidelity to the original data across all scales, as shown in \cref{fig:upsample_vis}.

\subsection{\modelname Architecture}

The \modelname architecture, depicted in \cref{fig:lg_arch}, begins by extracting \pixel features via a lightweight convolutional stem. Subsequently, SCA is employed to form \sp, significantly reducing local redundancies. This enables the application of ViT blocks directly on \sp, allowing the model to capture long-range dependencies and contextual information across superpixels.

To address the challenge of gradient conflicts, which arise from the simultaneous objectives of part and object segmentation \cite{wang2015joint}, \modelname incorporates two specialized branches designed to refine the segmentation process. In the part segmentation branch, \sp are further processed with additional ViT blocks, classified, and then upsampled to generate part segmentation predictions through our association-aware upsampling. Concurrently, for object segmentation, \sp are further abstracted by several ViT blocks, and undergo aggregation into \group via GCA. These \group are classified and upsampled to articulate the final object segmentation predictions, also utilizing the association-aware upsampling.
This dual-branch architecture optimally balances the demands of both segmentation tasks, ensuring accurate delineation of both parts and objects within a cohesive framework.

%% file: sec/4_experiments.tex
\section{Experiments}
\subsection{Datasets}
To benchmark \modelname, we evaluate on two benchmark datasets that include per-pixel part annotations: PartImageNet \cite{he2022partimagenet} and Pascal-Part \cite{chen2014detect}. PartImageNet augments 158 classes from the original ImageNet dataset with part annotations across 24,095 images. Pascal-Part is an enhancement of the VOC dataset \cite{Everingham2010ThePV} with 10,103 images across 20 classes. We focuses specifically on 16 classes with part-level annotations, following Compositor \cite{he2023compositor} protocols.

\subsection{Implementation Details}
\textbf{Hierarchical Feature Representation}\quad
\modelname delineates clear relationships among \pixel, \sp, and \group representations. The spatial dimensions of superpixel features are scaled down to a quarter of those of pixel features, while \group features further reduce to a $1/16$ of superpixel number, which are initialized by a $4\times4$ average pooling from \sp. This scaling strategy effectively balances detail retention and contextual abstraction. The model employs dual heads for \sp and six heads for \group in SCA and GCA operations, optimizing local-global contextual interactions.

\textbf{Attention Mechanism Integration}\quad
Cross-attention modules are integrated at strategic points within the ViT architecture, enhancing both shallow and deep layers. SCA blocks are positioned before the first and third self-attention layers, whereas GCA blocks are inserted ahead of the 9th, 10th, and 11th layers. We utilize the LayerScale technique \cite{touvron2021going} to promote stable training and faster convergence by ensuring uniform gradient distribution.

\textbf{Training Protocol}\quad
Our training configuration mirrors the parameters set by Compositor to facilitate direct comparisons. We employ AdamW \cite{loshchilov2019decoupled} with an initial learning rate of 0.0002, adjusting the ImageNet-pretrained backbone's learning rate to 10\% of this value. Learning rates decrease tenfold at 90\% and 95\% of the training timeline. Models undergo training for 50k iterations on PartImageNet and 10k on Pascal-Part, with a batch size of 128. Data augmentation techniques include random cropping and large-scale jittering \cite{du2021simple,ghiasi2021simple}.

\begin{table}[h!]
\setlength{\tabcolsep}{2pt}
\caption{Comparison of state-of-the-art methods on the PartImageNet and Pascal-Part validation splits.}
\begin{center}
\resizebox{0.9\textwidth}{!}{
\begin{tabular}{lccccccccccc}
\toprule
\multicolumn{1}{c}{\multirow{3}{*}{Method}} & \multicolumn{1}{c}{\multirow{3}{*}{Backbone}} & \multicolumn{1}{c}{\multirow{3}{*}{Params}} & \multicolumn{1}{c}{\multirow{3}{*}{Flops}} & \multicolumn{4}{c}{PartImageNet} & \multicolumn{4}{c}{Pascal-Part} \\
\cmidrule{5-8} \cmidrule{9-12}
& & & & \multicolumn{2}{c}{Part} & \multicolumn{2}{c}{Object} & \multicolumn{2}{c}{Part} & \multicolumn{2}{c}{Object} \\
\cmidrule{5-6} \cmidrule{7-8} \cmidrule{9-10} \cmidrule{11-12}
& & & & mIoU & mAcc & mIoU & mAcc & mIoU & mAcc & mIoU & mAcc \\
\midrule 
\textit{Separate Training} \\
\quad DeeplabV3+ \cite{chen2018encoderdecoder} & ResNet-50 \cite{he2015deep} & $  43\mathrm{M} \times 2 $ & $ 51\mathrm{G} \times 2 $ & $60.6 $ & $71.1 $ & $68.4 $ & $81.0 $ & - & - & - & - \\
\quad Maskformer \cite{cheng2021perpixel} & ResNet-50 \cite{he2015deep} & $  45\mathrm{M} \times 2$ & $ 53\mathrm{G} \times 2 $ & $60.3 $& $72.8 $ & $70.2 $& $82.0 $ & $47.6 $& $58.6 $ & $72.7 $& $81.9 $ \\
\quad SegFormer \cite{xie2021segformer} & MiT-B2 \cite{xie2021segformer} & $  24\mathrm{M} \times 2$ & $62\mathrm{G} \times 2$ & $62.0 $& $73.8 $ & $74.6 $ & $85.2$ & - & - & - & - \\
\quad Maskformer \cite{cheng2021perpixel} & Swin-T \cite{liu2021swin} & $  46\mathrm{M} \times 2$ & $ 55\mathrm{G} \times 2 $ & $64.0 $ & $77.4 $ & $77.9$ & $87.4$ & $55.4 $& $67.2 $ & $81.4 $& $89.3 $ \\
\quad \bf{\modelname}  & ViT-S \cite{dosovitskiy2021image}& $  34\mathrm{M} \times 2$ &  $ 50\mathrm{G} \times 2 $ &$ \mathbf{69.4}$ & $\mathbf{80.0}$&$ \mathbf{80.0}$ &$ \mathbf{89.3}$ &$ \mathbf{57.5}$ & $\mathbf{67.2}$ & $ \mathbf{85.1}$ &$ \mathbf{92.0}$ \\
\midrule 
\textit{Joint Training} \\
\quad Maskformer \cite{cheng2021perpixel} & ResNet-50 \cite{he2015deep} & $  50 \mathrm{M}$ & $ 53\mathrm{G} $ & $58.0 $& $70.4 $ & $70.4 $& $81.8 $ & $46.6 $& $58.0 $ & $72.1 $& $81.1 $\\
\quad Maskformer \cite{cheng2021perpixel} & Swin-T \cite{liu2021swin} & $  51 \mathrm{M}$ & $ 55\mathrm{G} $ & $61.7 $& $75.6 $ & $77.2 $& $87.1 $ & $54.2 $ & $66.4 $& $81.0 $& $88.7 $\\
\quad Compositor \cite{he2023compositor} & ResNet-50 \cite{he2015deep} & $  50 \mathrm{M}$ & 54G & 61.4 & 73.4 & 71.8 & 83.0 & $48.0 $& $58.8 $ & $74.4 $& $83.8$\\
\quad Compositor \cite{he2023compositor} & Swin-T \cite{liu2021swin} & $  51 \mathrm{M}$ & 57G & $64.6 $ &$ 78.3$ &$79.0$&$87.8$ & $55.9 $ &$67.6$ &$83.1$&$90.4$\\
\quad \bf{\modelname}  & ViT-S \cite{dosovitskiy2021image}& $  38 \mathrm{M}$ & 50G &$\mathbf{ 67.4}  $ & $\mathbf{79.6} $ &$\mathbf{79.8}$ & $\mathbf{88.4}$ &$ \mathbf{57.4}$ & $\mathbf{67.9}$ &$\mathbf{85.1}$ & $\mathbf{91.8}$ \\
\bottomrule
\end{tabular}
}
\end{center}
\label{tab:mainresult_combined}
\end{table}

\subsection{Main Results}
\label{sec:main_results}

Following the experimental setup established by Compositor \cite{he2023compositor}, we evaluated \modelname in both specialized and dual-task scenarios. As shown in \cref{tab:mainresult_combined}, \modelname, when jointly trained on dual tasks, achieves a part mIoU of 67.4\% and an object mIoU of 79.8\% on PartImageNet. These findings represent improvements over Compositor, with increases of 2.8\% and 0.8\% in part and object mIoU, respectively.
Further validation on Pascal-Part (\cref{tab:mainresult_combined}) corroborates these advancements, with \modelname achieving a part mIoU of 57.4\% and an object mIoU of 85.1\%, surpassing Compositor's performance by approximately 1.5\% and 2.0\%, respectively.

Considering Compositor's observation that dual-task frameworks might compromise individual task performance, we also investigated task-specific training for parts and objects, referred to as Separate Training in \cref{tab:mainresult_combined}. This approach significantly improved part mIoU on PartImageNet, highlighting \modelname's superior capability in precise part segmentation.

A crucial factor in \modelname's success is its optimized parameter use and computational efficiency. The model not only exceeds previous performance benchmarks but also does so with a lower total parameter count. This efficiency demonstrates the model's effectiveness in reducing redundancy across both local and global scales and enhancing accuracy through a sophisticated hierarchical semantic representation framework. The strategic balance between model complexity and performance underscores our approach's ability to refine segmentation outcomes without increasing computational demands.

\begin{figure}[t]
\centering
\begin{minipage}{0.42\textwidth}
    \centering
    \begin{subfigure}[b]{0.48\textwidth}
        \includegraphics[width=\textwidth]{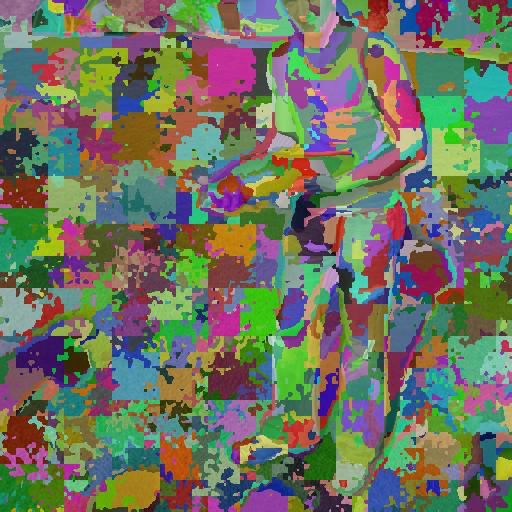}
        \caption{Part Semantics Emerge from Object Segmentation}
        \label{fig:sub1}
    \end{subfigure}
    \hfill
    \begin{subfigure}[b]{0.48\textwidth}
        \includegraphics[width=\textwidth]{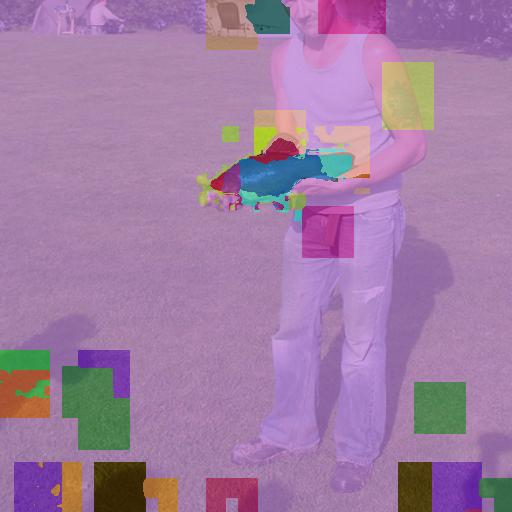}
        \caption{Object Semantics Emerge from Part Segmentation}
        \label{fig:sub3}
    \end{subfigure}
    \caption{
    \textbf{Visualization of Part and Object Semantic Emergence.} (a) In object segmentation, \sp reveal emerging part semantics. (b) Conversely, during part segmentation, object semantics become apparent within \group. 
        }
    \label{fig:emerge}
\end{minipage}%
\hfill
\begin{minipage}{0.55\textwidth}
    \centering
    \captionsetup{format=plain,labelsep=period,font=small,belowskip=-5mm}
    \includegraphics[width=\textwidth]{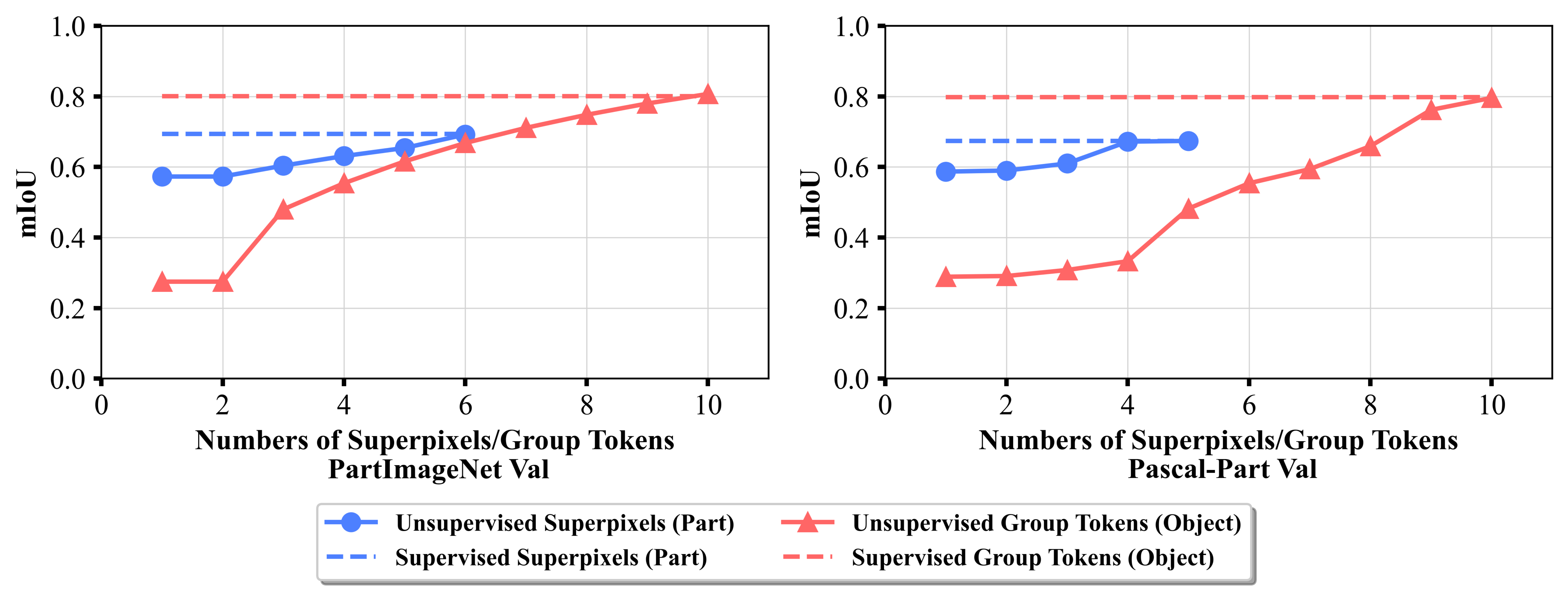}
    \caption{\textbf{Quantitive Evaluation of Unsupervised Superpixels and Group Tokens.}
    Employing 6 \sp for part segmentation and 10 \group for object segmentation in an unsupervised setting yielded mIoU scores comparable to those achieved with supervised methods.
    }
    \label{fig:oracle}
\end{minipage}
\end{figure}
\subsection{Semantic Hierarchy Emergence in Part and Object Segmentation}


Our model leverages a hierarchical design which mirrors natural segmentation layers of parts and objects. This structured approach prompts an examination of how \modelname's segments, specifically \sp and \group, handle semantic grouping under specific supervision scenarios. We investigate two key configurations: the capacity of \sp to cluster semantically in line with objects when guided solely by part annotations and the ability of object-level supervision on \group to implicit guide the semantic understanding within \sp. Such experiments concentrate on the object segmentation branch, as outlined in \cref{fig:lg_arch}, where supervision is deliberately limited to part or object annotations to distinctly observe the impact of hierarchical representations on semantic discovery.

\textbf{Qualitative Emergence Evaluation.}
The semantic arrangement of \sp and \group is visually assessed by identifying the most relevant entities through the argmax across association matrices, as illustrated in \cref{fig:emerge}. This approach reveals spontaneous semantic emergence: part-only supervision leads to object-level semantic recognition within \group. Inversely, object-only supervision enables the delineation of part semantics within \sp, effectively identifying parts as class-specific patterns \cite{tang2017learning}. Remarkably, both \sp and \group show consistent alignment with the physical boundaries of parts or objects, even with a straightforward argmax selection. This alignment indicates that the hierarchical structuring effectively maintains detailed spatial information, which is critical in precise segmentation tasks.

\textbf{Quantitative Emergence Evaluation.} Complementing our qualitative analysis, a quantitative assessment further validates the emergence phenomena observed. We utilize an oracle setup where the model is trained with object-level annotations but evaluated \sp against part-level ground truths, and vice versa for \group. Here, we employ the Mask-to-Attention conversion method from SegViT \cite{zhang2022segvit} to derive segmentation masks used for mIoU calculations. Remarkably, selecting a small subset of top-k \sp or \group for evaluation yields mIoU scores that are on par with those obtained in fully supervised settings, as shown in \cref{fig:oracle}. This experiment not only supports the model's ability to mimic aspects of human visual processing with minimal supervision but also showcases its efficiency in managing redundancies across various scales.

In summary, the evaluations both qualitative and quantitative, firmly establish that \sp and \group are capable of semantic emergence without direct supervision. This success underscores the effectiveness of our hierarchical design in simulating the nuanced processes of the human visual system, as it categorizes and assimilates visual information into coherent entities.

\subsection{Robustness to Occlusion}

\begin{figure}[htbp]
\centering
\includegraphics[width=0.85\textwidth]{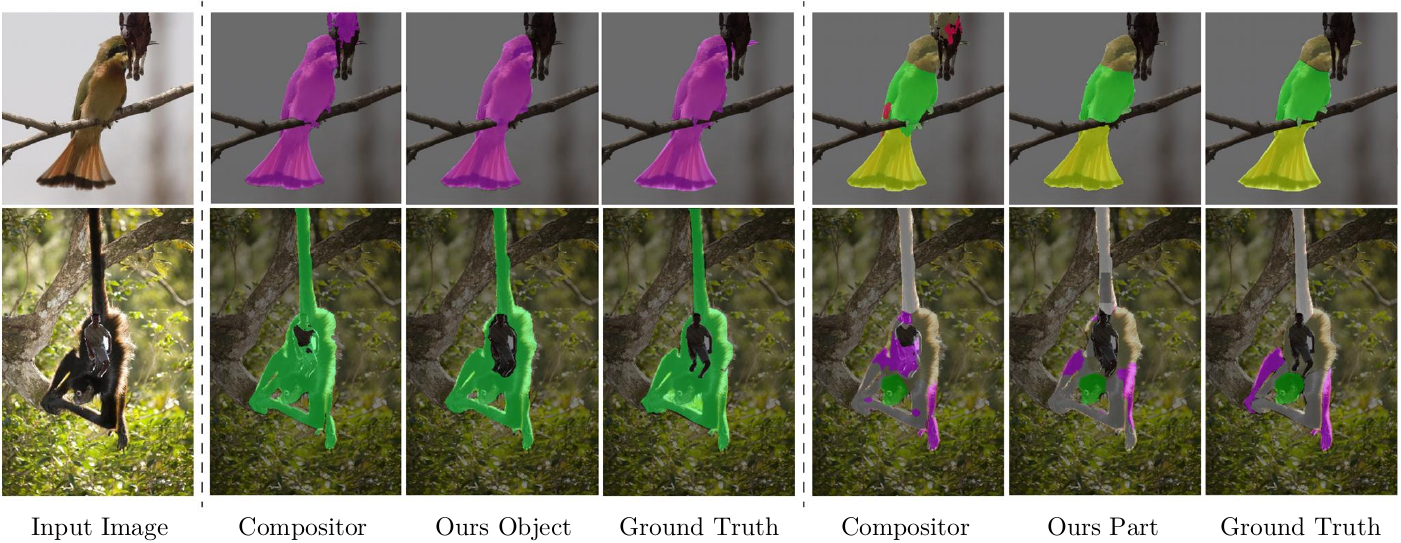}
\caption{Qualitative evaluation of images with occlusions.}
\label{fig:occlusion_result}
\end{figure}

To further test the robustness of \modelname, we evaluate its performance on the Occluded-PartImageNet-v1 dataset \cite{he2023compositor}, where 20\%-40\% of the object region is obscured. Given the occlusions, \modelname's performance drops 8.0\% in part mIoU and 15.5\% in object mIoU — positioning it favorably against benchmarks set by MaskFormer and Compositor. The qualitative results of this evaluation, illustrated in \cref{fig:occlusion_result}, further confirm \modelname's capability to robustly handle segmentations even in scenarios involving significant occlusions, reflecting its practicality for real-world applications.


\begin{table}[htbp]
\centering
\footnotesize
\caption{Qualitative results on Occluded-PartImageNet-v1. \modelname shows a smaller performance drop on occluded images compared to MaskFormer and Compositor.}
\setlength{\tabcolsep}{14pt} 
\begin{tabular}{ccc}
   \toprule
   Method  & Part mIoU & Object mIoU  \\
   \midrule
   MaskFormer  & 50.2 \scriptsize{(-13.7)} & 56.7 \scriptsize{(-21.2)} \\
   Compositor & \textbf{54.6} \scriptsize{(-10.0)} & \textbf{63.7} \scriptsize{(-15.2)} \\
   \textbf{LGFormer(Ours)}  & \textbf{59.3} \scriptsize{(-8.1)} & \textbf{64.2} \scriptsize{(-15.6)} \\
   \bottomrule
\end{tabular}
\label{tab:occlusion_results}
\end{table}

\subsection{Ablation Study}

\begin{table}[htbp]

\begin{center}
\resizebox{0.8\textwidth}{!}{
\setlength{\tabcolsep}{1pt} 
\begin{tabular}{lccccc}
\toprule
\multicolumn{1}{c}{\multirow{2}{*}{Method}} & \multicolumn{1}{c}{\multirow{2}{*}{\#Params}} & \multicolumn{2}{c}{Part} & \multicolumn{2}{c}{Object}  \\
\cmidrule{3-4} \cmidrule{5-6}
& & mIoU & mAcc & mIoU & mAcc \\

\midrule
\textit{Number of group tokens} \\
\quad 64 group & $39 \mathrm{M}$ & 67.4 & 79.6 & 79.8 & 88.4  \\
\quad 256 group & $39 \mathrm{M}$ & 67.2  & 79.0 & 78.2 & 86.7 \\
\quad 16 group & $39 \mathrm{M}$ & 67.3 & 79.1 &78.6 & 87.2 \\

\midrule

\textit{Group token initialization method} \\
\quad avgpooling & $39 \mathrm{M}$  & 67.4 & 79.6 & 79.8 & 88.4  \\
\quad learnable & $39 \mathrm{M}$ & 67.1 & 79.0 &79.3 & 87.4\\
\quad conv & $41 \mathrm{M}$ &66.7 &78.2 &79.1 & 88.0 \\

\midrule

\textit{Number of GCA stages} \\
\quad 3 stages & $39 \mathrm{M}$ & 67.4 & 79.6 & 79.8 & 88.4  \\
\quad 4 stages & $40 \mathrm{M}$ & 67.0 & 78.9 & 78.5 & 87.3 \\
\quad 2 stages & $37 \mathrm{M}$ & 67.1 & 79.0 & 77.7 &  86.4 \\

\midrule

\textit{Upsampling Method} \\
\quad  Association-Aware Upsampling & $39 \mathrm{M}$ & 67.4 & 79.6 & 79.8 & 88.4  \\
\quad  Bilinear Upsampling & $39 \mathrm{M}$ & 65.3 & 76.8 & 73.2 & 82.8  \\

\bottomrule
\end{tabular}
}
\end{center}
\caption{Ablation Study on PartImageNet val split.}
\label{tab:ablation}
\end{table}

In our ablation study, we systematically explore the design choices of \modelname to validate the configuration's impact on segmentation performance. The results, detailed in \cref{tab:ablation}, affirm the efficacy of our methodological choices by highlighting the role of group token quantity, branch block optimization, and initial group token methods.

\textbf{Optimal Group Token Count.}  Adjusting the number of \group tokens demonstrates the critical balance required for precise semantic detail capture within hierarchical aggregation strategies. Reducing \group tokens to 64 or increasing to 256 affects object mIoU negatively by 1.6\% and 1.2\%, respectively. This phenomenon underscores that while fewer \group tokens (16) maintain robust performance at sparse resolutions ($4 \times 4$), excessive quantities may dilute semantic richness and increase computational overhead. Our optimal count demonstrates how the model efficiently condenses semantic information without losing detail or explainability.

\textbf{Group Token Initialization.} Our comparative analysis highlights the superiority of average pooling over learnable tokens or convolution-based methods for initializing group tokens. This simplicity aligns with our methodological emphasis on efficiency, corroborating our claim that average pooling adequately prepares group tokens for subsequent hierarchical processing without necessitating complex initialization techniques.

\textbf{Branch Block Configuration.} The strategic arrangement of branch blocks within \modelname is critical for dealing with gradient conflict and ensuring model capacity. Reducing branch blocks from three to two, or increasing them to four, adversely affects part and object segmentation mIoU by 0.3\% and 2.1\%, and 0.4\% and 1.3\%, respectively. These outcomes validate our choice of employing three branch blocks as the optimal configuration, effectively balancing effective gradient management with the preservation of hierarchical modeling capabilities.

\textbf{Upsampling Method.} The \upsamplename method surpasses traditional bilinear upsampling, enhancing part mIoU by 2.1\% and object mIoU by 6.6\%. The significant improvement in object mIoU can be attributed to the inherent limitations in bilinear upsampling, particularly its inability to recover substantial edge details lost when images are downsampled by 32. Our \upsamplename method progressively reconstructs details at the part and pixel levels, preserving essential information to refine prediction accuracy. This detailed recovery process significantly boosts segmentation performance, showcasing the method's capability to retain and reconstruct fine details for enhanced part and object segmentation outcomes.

%% file: sec/5_conclusion.tex
%

\section{Conclusion}


In this paper, we introduce \modelname, a hierarchical transformer-based model for advanced image segmentation, bridging the granularity of part segmentation with the comprehensive scope of object segmentation. Our multi-level representation strategy progresses from pixels to superpixels and finally to cohesive groups, supported by local and global aggregation strategies. Local aggregation forms superpixels aligned with object parts, while global aggregation organizes these superpixels into larger groups corresponding to entire objects. This dual framework ensures adaptability to various supervision inputs while maintaining computational efficiency and enhancing the segmentation performance.
